%% file: main.tex
\documentclass[10pt,table,twocolumn,letterpaper]{article}

\usepackage[pagenumbers]{cvpr} 

\definecolor{lightorange}{RGB}{250, 237, 205}
\definecolor{lightblue}{RGB}{202, 240, 248}
\definecolor{lightgreen}{RGB}{233, 237, 201}

\newcommand{\ours}{\textit{MMFactory}}

\definecolor{cvprblue}{rgb}{0.21,0.49,0.74}
\usepackage[pagebackref,breaklinks,colorlinks,allcolors=cvprblue]{hyperref}
\usepackage{multirow}
\usepackage{array}
\PassOptionsToPackage{table,xcdraw}{xcolor}
\usepackage{tabularx}
\usepackage{subcaption}
\usepackage{multicol}

\def\paperID{6585} 
\def\confName{CVPR}
\def\confYear{2025}

\title{MMFactory: A Universal Solution Search Engine for Vision-Language Tasks}

\author{Wan-Cyuan Fan$^{1,2}$ \ \ \ \  Tanzila Rahman$^{1,2}$ \ \ \ \  Leonid Sigal$^{1,2,3}$\\
$^1$University of British Columbia \ \ \ \ $^2$Vector Institute for AI \ \ \ \ $^3$CIFAR AI Chair\\
{\tt\small \{wancyuan, trahman8, lsigal\}@cs.ubc.ca}
}

\begin{document}

\maketitle
\input{sec/0_abstract}    
\input{sec/1_intro}
\input{sec/2_related}
\input{sec/3_method}

\input{sec/4_exp}
\input{sec/5_conculsion}

{
    \small
    \bibliographystyle{ieeenat_fullname}
    \bibliography{main}
}



\end{document}

%% file: sec/0_abstract.tex
\begin{abstract}

With advances in foundational and vision-language (VLM) models, and effective fine-tuning techniques, a large number of both general and special-purpose models have been developed for a variety of visual tasks. 
Despite the flexibility and accessibility of these models, no single model is able to handle all tasks and/or applications that may be envisioned by potential users. Recent approaches, such as visual programming and multimodal LLMs with integrated tools aim to tackle complex visual tasks, by way of program synthesis. However, such approaches overlook user constraints (e.g., performance / computational needs), produce test-time sample-specific  solutions that are difficult to deploy, and, sometimes, require low-level instructions (e.g., code snippets for similar problems) that maybe beyond the abilities of a naive user. To address these limitations, we introduce \ours, a universal framework that includes model and metrics routing components, acting like a solution search engine across various available models. Based on a task description and few sample input-output pairs and (optionally) resource and/or performance constraints, \ours~can suggest a diverse pool of programmatic solutions by instantiating and combining visio-lingual tools (e.g., detection, segmentation, VLMs) from its model repository. In addition to synthesizing these solutions, \ours~also proposes metrics and benchmarks performance / resource characteristics, allowing users to pick a solution that meets their unique design constraints. From the technical perspective, we also introduced a committee-based solution proposer that leverages multi-agent LLM conversation to generate executable, diverse, universal, and robust solutions for the user. Experimental results show that \ours~outperforms existing methods by delivering state-of-the-art solutions tailored to user problem specifications. 
\end{abstract}

\vspace{-0.4in}

%% file: sec/1_intro.tex
\section{Introduction}
\label{sec:intro}
 
Large language models (LLMs), such as GPT~\cite{achiam2023gpt} and Gemini~\cite{team2023gemini,team2024gemini}, have demonstrated powerful capabilities across various domains, significantly transforming how people approach their tasks and even their daily lives. Building on these models, a wide range of vision-language (VLM) or multimodal LLMs (MLLMs)~\cite{liu2023llava, zhang2023internlm, bai2023qwen, awadalla2023openflamingo, abdin2024phi} have been developed by integrating modality adapters and encoders into their frameworks. 
This advancement has resulted in state-of-the-art models capable of solving complex visual tasks.

\input{figures/teaser}


Despite the push for building AGI-like agents, that are all capable, even models like GPT-4o tend to be inferior, or lacking, on specific tasks~\cite{fu2024blink, li2024seed}. At the same time, 
with the development of fine-tuning techniques, customized or expert models tailored to specific tasks have become easier to develop. 
With different training data, fine-tuning approaches, and frameworks, models with varying specialties and characteristics are being introduced daily. 
One can imagine that in near future such models will be ubiquitous, creating a marketplace of agents with an overwhelming design choices for users to pick from and build on. 
In this scenario, routing approaches are needed that can take user-defined tasks, needs, and constraints, acting as a search engine among all types of models, to provide suggested solutions for the user.

Previous works in visual programming~\cite{suris2023vipergpt, gupta2023visual} and multimodal language models (MLLMs) with tool  integration~\cite{qin2023toolllm, hu2024visual, lu2024chameleon} have explored using LLMs as planners to utilize external tools or APIs for solving complex visual tasks or to decompose tasks into sub-tasks. While these approaches have shown promise, there are several limitations to consider. 
First, existing methods assume a single specialized tool for a given sub-task ({\em e.g.}, detection \cite{liu2023grounding}, segmentation \cite{kirillov2023segment}, depth estimation \cite{yang2024depth}). 
This is overly simplistic, as a variety of tools exist for any one sub-task, inculcating within a particular family of models, that differ by backbone, number of parameters and overall performance. 
Second, these works generally overlook the user’s specific computation needs and constraints when generating solutions, resulting in inability to tailor solutions to particular hardware or deployment cost ({\em e.g.}, a user maybe willing to forgo 1\% better performance if inference cost is reduced by 50\%). 
Third, the proposed solutions are often tailored per specific example or scenario, which limits their generalization and applicability to other examples in the task, as shown in~\cref{fig:teaser}. 
Deployment of such solutions is problematic ({\em e.g.}, no constant code path exist that maybe distilled to a small model executable on an edge device). 
Addressing these limitations is essential for creating more versatile and user-centric framework for routing the solutions among different kinds of models in order to create custom agents capable of solving specific user problems in accordance to their specification. 

To address these challenges, in this work, we introduce \ours~-- a universal framework for automatic and programmatic development of task-specific agents. \ours~(Fig.~\ref{fig:teaser}a) includes a model and metric routing components; that, in combination, act as a solution search engine for non-expert users. 
Based on a task description ({\em e.g.}, comparison of depth of points in an image), a few sample input-output pairs ({\em e.g.}, set of images with labeled points and which point is closest to camera in each), and (optionally) resource and/or performance constraints ({\em e.g.}, compute limit), \ours~can suggest a diverse pool of programmatic solutions by instantiating and combining visual, LLM and VLM tools from its repository. 
In addition to synthesizing these solutions, \ours~also proposes metrics and benchmarks performance / resource characteristics, allowing users to pick a solution that meets their unique design constraints. From the technical perspective, we also introduced a committee-based solution proposer that leverages multi-agent LLM conversation to generate executable, diverse, universal, and robust solutions for the user. 

Notably, unpublished and concurrent work of \cite{ong2024routellm} also explores the idea of routing, but mainly for choosing a single (most accurate) among the $K$ possible LLM / VLM models (see~\cref{fig:teaser}c). \ours~framework is considerably more general and provides user with family of solutions and their performance characterization. In addition, our solutions, similar to visual programming \cite{suris2023vipergpt, gupta2023visual}, are drawn from an exponential set of tools that can work in tandem with one another. Further, the fact that our framework proposes solutions that contain a single executable code path, makes them much easier to deploy.


Our contributions are multiple fold. First, to the best of our knowledge, this work is the first to explore routing across vision, language, and vision-language models. Second, our propose framework can provide multiple solutions in a solution pool for user-defined tasks and constraints. Third, we introduce a novel approach that combines routing and a multi-agent solution proposer to deliver robust results. Fourth, unlike existing approaches, our proposed framework solves all instances of a user-defined task collectively, rather than generating separate solutions for each instance. Fifth, experiments on two benchmarks demonstrate that our framework outperforms the state-of-the-art.








%% file: figures/teaser.tex
\begin{figure}[t]
  \centering
  \includegraphics[page=2, trim={170 20 160 10}, clip, width=\columnwidth]{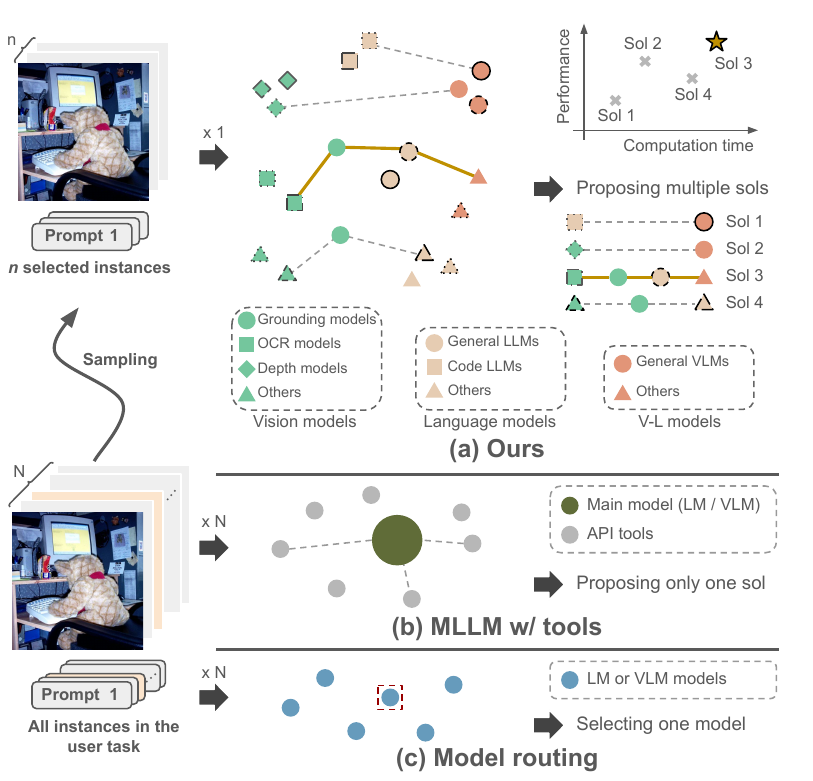}
  \vspace{-5mm}
  \caption{\textbf{Illustration of MMFactory}. Proposed MMFactory framework (a) contrasted with model routing approaches (c) and multimodal LLM with tools (b). Unlike  both prior classes of methods, MMFactory proposes a pool of programmatic solutions, composed of series of selected models from the pool, for a given task while also benchmarking their performance and computational characteristics. See Section~\ref{sec:intro} for full discussion.}
  \label{fig:teaser}
  \vspace{-4mm}
\end{figure}

%% file: sec/2_related.tex
\vspace{-0.05in}
\section{Related works}
\label{sec:related_works}


\noindent
{\bf Multimodal Large Language Models.} 
Building on the recent success of large language models (LLMs)~\citep{gpt4, gemini, claude3, touvron2023llama}, research trends have shifted toward enhancing these LLMs with multi-modal capabilities. Some of these MLLMs~\cite{liu2024visual,zhang2023internlm, lu2024deepseek, mckinzie2024mm1} are created for general purpose, while others are designed for specific tasks, including coding~\cite{roziere2023code,githubcopilot,jiang2023mistral}, video understanding~\cite{zhang2023video, chen2023videollm}, 3D~\cite{zhu20233d,hong20233d,chen2024spatialvlm}, audio or speech~\cite{fathullah2024prompting, das2024speechverse, borsos2023audiolm}, math~\citep{azerbayev2023llemma,wang2023mathcoder}, scientific chart~\citep{fan2024pre,han2023chartllama,meng2024chartassisstant}, and robotics~\cite{zeng2023large, brohan2023rt}, which has demonstrated promising results. However, 
language-based models alone can’t handle complex tasks very well. Multimodal models, which combine text and images, also face challenges, like misinterpreting context when information is split across text and visuals. They might connect unrelated details or miss important clues, leading to errors. Therefore, researchers are exploring tools and interactive systems to improve their understanding of multimodal information.

\input{figures/framework}


\vspace{0.1in}
\noindent
{\bf Visual programming, LLMs with tools and Routing.}
As humans, when we face complex tasks, we decompose them into subtasks to understand them better or use tools to make them simpler.
These concepts have been extended to neural networks, where previous works~\cite{andreas2016neural, johnson2017inferring} suggest that complex vision tasks are fundamentally compositional and can be divided into atomic perceptual units. Following this concept, visual programming~\cite{gupta2023visual, suris2023vipergpt} and LLMs with tool~\cite{liu2023llava, patil2023gorilla, qin2023toolllm} become prominent research trends. Practically, visual programming focus on leveraging LLMs' coding ability to decompose complex tasks into multi-step Python code with specialized vision tools. On the other hand, LLMs with tool use focus on teaching LLMs to use various types of tools to achieve image generation/editing~\cite{liu2023llava}, accessing web engines~\cite{patil2023gorilla, lu2024chameleon}, operating systems~\cite{packer2023memgpt}, {\em etc}. However, these methods have a common problem that the multimodal modules are designed for specific tasks and can’t be reused for similar ones. They also don’t consider user constraints like model size, complexity, or preferences. This gave rise to routing-based approaches. In these approaches~\cite{ong2024routellm, hu2024routerbench, shnitzer2023large}, a router model can switch between a stronger or weaker LLM during inference to balance cost with model performance. However, this method still needs the router LLM to be trained and can’t offer versatile solutions based on user needs. It also relies on a single model, which isn’t enough to solve a complex task efficiently. In contrast, our framework provides multiple options (\emph{i.e.} solution pool) for users to choose from, and these options are versatile and can be reused across all instances of the task, rather than being limited to individual instance.



\vspace{0.05in}
\noindent
{\bf From multi-modal Agents to multi-agent frameworks.}
Recently, due to the powerful reasoning, tool usage, and other capabilities of LLMs, these models have become essential building blocks in the development of artificial intelligence agents~\cite{wu2023autogen, Significant_Gravitas_AutoGPT, Chase_LangChain_2022, li2023camel} for many real-world applications, such as medicine~\cite{tang2023medagents}, general tasks~\cite{chen2023autoagents}, and robotics~\cite{kannan2023smart}. Given the increasing complexity of tasks, an intuitive approach is to enhance the capabilities of agents by incorporating multiple agents into the task solving. Previous works have showcased that multi-agents conversations or debates can improve various capabilities, such as divergent thinking~\cite{liang2023encouraging}, factuality and reasoning abilities~\cite{du2023improving}, and validation~\cite{wu2023empirical}, and can even achieve automatic agent creation~\cite{chen2023autoagents}. Among these, the most relevant to our work is 
AutoAgents~\cite{chen2023autoagents}, which introduces ``observers" to monitor multi-agent conversations, helping ensure quality and coherence in responses. However, AutoAgents provides only one solution per prompt, while our approach offers multiple options with performance and cost details to help users choose the best fit. Additionally, AutoAgents relies on GPT-4's reasoning, limiting its flexibility with open-source models and restricting it to tasks like open-ended questions and creative writing. Our system, in contrast, supports any open-source model and can handle a wide range of vision tasks. Most importantly, AutoAgents focus on dynamically creating multiple agents based on the task content and planning solutions. Our approach, however, focuses on solving tasks by routing different vision models and incorporating a Python coding environment, which AutoAgents have not explored.

%% file: figures/framework.tex
\begin{figure*}[t]
  \centering
  \includegraphics[page=3, trim={0 60 5 60}, clip, width=0.9\textwidth]{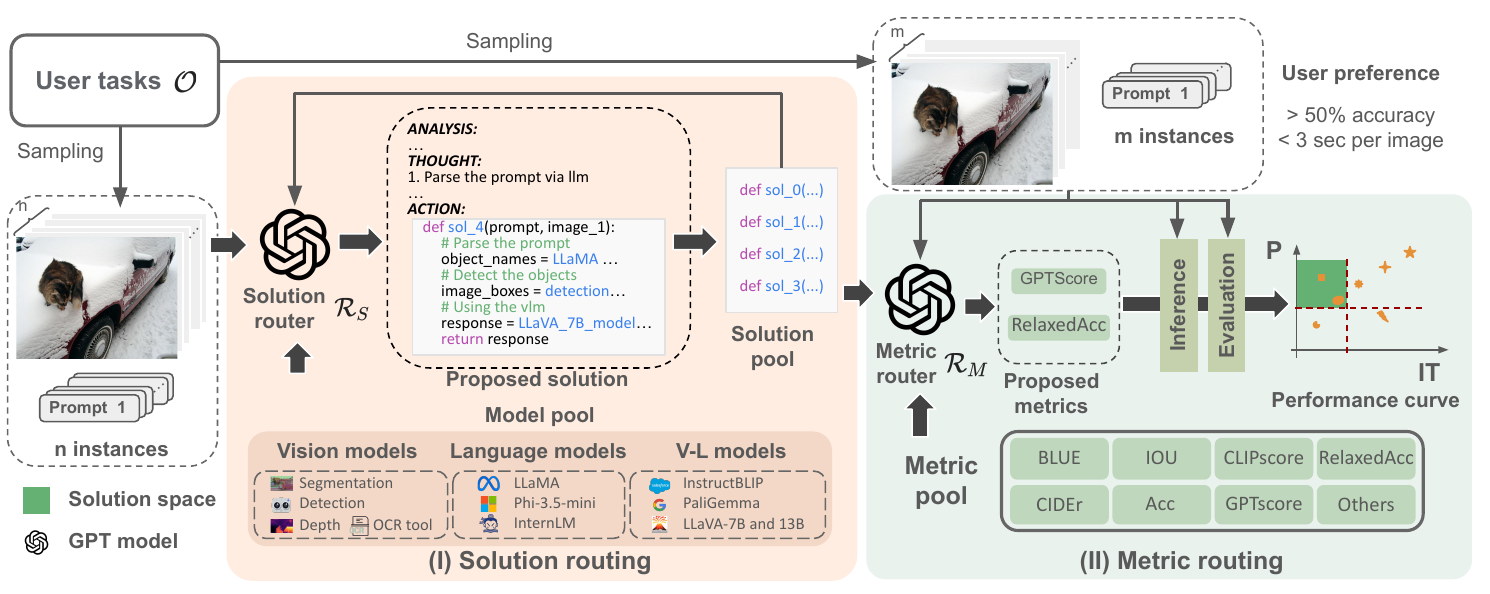}
  \vspace{-4mm}
  \caption{\textbf{Overview of \ours.} Our framework includes two primary components: Solution Router and  Metric Router. The Solution Router generates a pool of potential solutions for the task, while the Metric Router evaluates these solutions, estimating their performance and computational cost to generate a performance curve. This curve enables users to select the model optimal for their task requirements.}
  \label{fig:framework}
  \vspace{-4mm}
\end{figure*}

%% file: sec/3_method.tex
\section{Methodology}
\label{sec:method}

\input{figures/demo_inputs}

\subsection{Overview of MMFactory}
We introduce \ours, an universal framework designed not only to propose programmatic solutions based on user-defined task examples but also to provide estimated performance and time costs for each solution, allowing users to make informed choices. This framework functions like a solution search engine and interface across various models, enabling access to models for task-solving without requiring extensive background knowledge. 
\ours~has several unique features. In addition to proposing multiple solutions with estimated performance and cost plots, the solutions generated are general and can be applied across all examples within the specified task. Specifically, \ours ~consists of two key components: the {\em Solution Router} and {\em Metric Router}. The former can generate multiple general solutions for solving the task, while the later evaluate the solutions to estimate their performance and computation cost. The framework is illustrated in~\cref{fig:framework}.\footnote{Our entire framework is built using Autogen~\cite{wu2023autogen}, an open-source programming framework for agentic AI design that enables the development of multi-agent communication and Python code execution environments.}
Furthermore, we leverage advanced multimodal LLMs ({\em e.g.}, GPT) as the solution and metric routers.
For better understanding, we first introduce the necessary notations, followed by a detailed explanation of these two modules in the following sections. 

\vspace{0.1in}
\noindent
{\bf Problem Formulation and Notations.}
As shown in~\cref{fig:framework}, given a user-specified task with $N$ instances, we represent these instances as a set $\mathcal{O} = \{\hat{o}_1, \hat{o}_2, \dots, \hat{o}_n, o_{n+1}, \dots, o_N\}$, where $\hat{o}_i = (\mathcal{I}_i, q_i, a_i)$ and $o_i = (\mathcal{I}_i, q_i)$, with $\mathcal{I}_i$, $q_i$, and $a_i$ denoting the image set, task request prompt, and ground-truth answer for that instance, respectively. Note that only $n$ instances have ground-truth answers, referred to as example instances $\mathcal{O}_{\text{ex}} = \{\hat{o}_1, \hat{o}_2, \dots, \hat{o}_n\}$, where $n \ll N$. 
The goal of the Solution Router, $\mathcal{R}_S$, is to propose programmatic solutions for the task based on the example instances so that the answers for all instances can be inferred by leveraging the proposed solutions. In practice, together with the example instances and predefined task-agnostic prompts (e.g., model definitions), we construct an input prompt $\mathcal{P}$ for $\mathcal{R}_S$ to generate a solution pool $\mathcal{S} = \{s_1, s_2, \dots, s_l\}$.
Note that we set $n \ll N$ to enable model routing to perform reasoning to obtain the answer rather than simply memorizing the ground truth answers. 
Once solutions are obtained, the Metric Router, $\mathcal{R}_M$, samples a subset with $m$ instances from $\mathcal{O}$ to evaluate the performance of each solution in $\mathcal{S}$. This evaluation yields a set $\mathcal{E} = \{(p_1, c_1), (p_2, c_2), \dots, (p_l, c_l)\}$, where $p_i$ and $c_i$ denote the performance and computation cost of the $i$-th solution in $\mathcal{S}$. Optionally, other metrics can also be logged.

\subsection{Inputs Structure for Solution Router}
As mentioned in the previous section, our framework is designed to propose multiple solutions that leverage models in the model pool to solve the task. The challenging aspects of this task is that the router must not only understand the task but also comprehend the details of each model in the pool to ensure correct use in the solution. For such complex task, in addition to the initial task prompt, we have to provide extra details for the router, including definitions of the models in the model pool, a requirements list, in-context examples, and the solution pool. For each task, the input prompt $\mathcal{P}$ structure is detailed below (examples can also be found in the supplementary.) consisting of \textit{task-agnostic information}:

\begin{itemize}
    \item \textbf{Model definitions $\mathcal{P}^{d}$}: Describes the details of each model in the model pool, including functionality, input arguments, return arguments, and example use cases.
    
    \item \textbf{Requirements $\mathcal{P}^{r}$}: A predefined list of requirements for the router to consider when generating solutions.
    
    \item \textbf{In-context examples $\mathcal{P}^{e}$}: Following previous work~\cite{hu2024visual}, we provide four different output examples as references. Note that the in-context examples are not sampled from the user task $\mathcal{O}$. 
    
    \item \textbf{Solution pool $\mathcal{P}^{s}$}: Showcases all previously generated solutions (Python code only). If no solution exists, "EMPTY" will be displayed.
\end{itemize}

\noindent
and \textit{user-specified task-specific instructions}:

\begin{itemize}
    \item \textbf{User-specification $\mathcal{P}^{u}$}: Contains the task definition, example instances $\mathcal{O}_{\text{ex}}$ sampled from the target task, and (optional) user constraints. Note that the task instances' input includes images.
    User input is illustrated in~\cref{fig:demo_inputs}.
\end{itemize}

\subsection{Multi-agent solution router}
Taking all the aforementioned information as input, the goal of solution router is to propose novel solutions to solve the task at hand. To achieve that, inspired by multi-agent conversation works~\cite{du2023improving}, we deploy multi-agent system for this complex 
problem. A conversation is instantiated between two teams:  
the solution proposer team and the committee team. The proposer team generates ideas and solutions, while the committee team checks for correctness, redundancy, and alignment with requirements, providing feedback. Each team consists of members and a leader. After gathering responses from their members, the leaders of two team exchange responses and collect feedback. By iteratively refining the solution based on this feedback, we achieve robust results. An illustration is provided in~\cref{fig:framework_agent}. We now detail each component and the conversation process within the multi-agent system. Please refer the supplement for example responses from all the agents in the solution Router.

\input{figures/multiagent}


\vspace{0.1in}
\noindent
{\bf Solution Proposing Team.} 
The solution proposing process involves three key components: (i) analyzing existing solutions and committee feedbacks, (ii) outlining step-by-step high-level instructions, and (iii) developing Python code implementation. This process integrates analysis, creative problem-solving, and rigorous coding. We employ two agents for this purpose: the solution proposer $\mathcal{A}_{sp}$ and the solution engineer $\mathcal{A}_{se}$ (see Fig.~\ref{fig:framework_agent}). 
The solution proposer $\mathcal{A}_{sp}$ begins by reviewing existing solutions and generating a novel approach with clear, high-level instructions, resulting in the ANALYSIS and THOUGHT sections of the output. Following this, the solution engineer $\mathcal{A}_{se}$ builds on the instructions provided to produce executable Python code, documented in the ACTION section. Together, the ANALYSIS, THOUGHT, and ACTION sections form a comprehensive solution for further review. Please refer to the supplementary materials for output examples.


\vspace{0.1in}
\noindent
{\bf Solution Committee Team.}
The Solution Committee oversees the quality and robustness of the generated solutions. Its main objectives are to verify that each solution meets predefined requirements, ensure code correctness and functionality, and check for redundancy with existing solutions. A significant challenge is validating code logic beyond mere error-free execution. 
Therefore, we introduce a code debugger that analyzes intermediate results. Additionally, with the code executor, we can provide the committee with intermediate outputs, enabling a detailed, step-by-step review of the logic. 
As shown in~\cref{fig:framework_agent}, we introduce two additional agents with specific roles: a requirement checker and a code checker. The requirement checker evaluates whether the solution aligns with the specified requirements. Meanwhile, the code checker assesses both intermediate and final execution results to verify the accuracy and logical soundness of the code. In the final stage, the repetition checker ensures that the proposed solution doesn’t duplicate any existing solutions in the current solution pool. If the logic of proposed solution matches an existing one, it rejects the solution to avoid redundancy in the solution pool.
Please refer to the supplemental for output examples.

\vspace{0.1in}
\noindent
{\bf Conversation between solution proposer and committee.}
The interaction between the Solution Proposing and Solution Committee Teams refines solutions iteratively, as depicted in~\cref{fig:framework_agent}. As mentioned in the prior work~\cite{du2023improving}, a multi-agent conversation framework enhances reasoning and improves solution accuracy. However, excessive iterations can lead to error propagation. To address this, we require each committee member to deliver a decision at every iteration, either accepting or rejecting the solution with feedback. If all committee members accept the solution, the iteration concludes. Recognizing that convergence is sometimes challenging, we enforce a maximum number of iterations. At the end of the conversation, if the final solution is not redundant (as confirmed by the repetition checker), the most recent iteration’s solution is preserved.

\input{tables/main}
\subsection{Metric Router}
After model routing, we are able to collect a pool of diverse solutions, $\mathcal{S} = \{s_1, s_2, \dots, s_m\}$ (see~\cref{fig:framework}). The evaluation router further assesses these solutions, resulting in a set $\mathcal{E} = \{(p_1, c_1), (p_2, c_2), \dots, (p_m, c_m)\}$, where $p_i$ and $c_i$ represent the performance and computation cost of the $i$-th solution in $\mathcal{S}$. We introduce an evaluation router, similar to the solution router, which uses the multimodal LLM’s reasoning to select the right metric based on the user’s task and the format of ground truth and predictions. Once the metric is chosen, we can proceed with performance testing and evaluation, estimating both the performance and cost of each solution.
The user can also supply a custom metric rendering evaluation router unnecessary; however, the choice of the metric may not itself be trivial for a naive user.

\vspace{0.1in}
\noindent
{\bf Input Structure.}
We again use MLLM ({\em i.e.}, GPT-4) as the router to select metrics for evaluation. Below, we detail the input prompt for the router, comprising of \textit{task-agnostic}:

\begin{itemize}
    \item \textbf{Metric Definitions}: Provides details for each metric in the metric pool, including use cases, input arguments, return arguments, and examples.
    
\end{itemize}

\noindent
and \textit{user-derived task-specific} instructions:

\begin{itemize}
    \item \textbf{Task Instances}: Similar to the solution router, this includes task instructions and $n$ example instances sampled from the target task, along with ground truth answers and predictions from the solutions.
\end{itemize}

\vspace{0.1in}
\noindent
{\bf Performance and Computation Cost Curve.}
For each proposed solution, we apply the aforementioned metric routing. Once a metric is selected, we first choose larger test cases from the user-provided task. As shown in~\cref{fig:framework}, we then perform evaluations, recording both performance and computation cost, and generate a plot. This allows users to select solutions based on their preferences. Please see supplemental for further discussion on metric routing.

%% file: figures/demo_inputs.tex
\begin{figure}[t]
  \centering
  \includegraphics[page=19, trim={190 70 190 40}, clip, width=0.90\columnwidth]{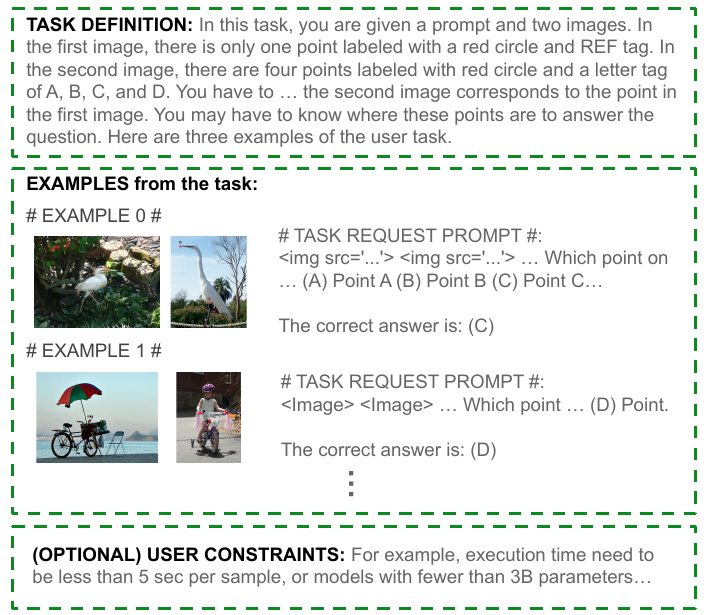}
  \vspace{-3mm}
  \caption{\textbf{Illustration of user specification inputs} $\mathcal{P}^u$.}
  \label{fig:demo_inputs}
  \vspace{-5mm}
\end{figure}

%% file: figures/multiagent.tex
\begin{figure}[t]
  \centering
  \includegraphics[page=5, trim={185 40 210 40}, clip, width=0.85\columnwidth]{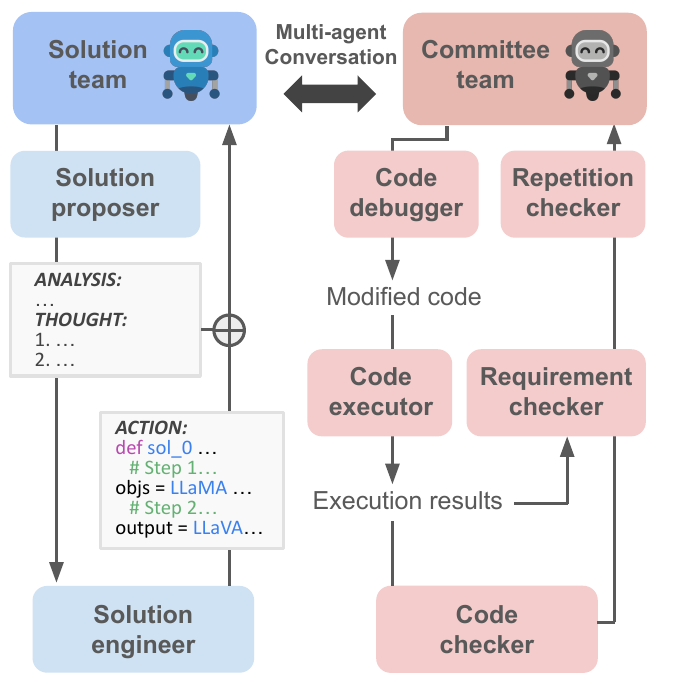}
  \vspace{-3mm}
  \caption{\textbf{Illustration of multi-agent conversation}. In the solution router, we have two team of agents performing conversation to get the final outputs.}
  \vspace{-0.15in}
  \label{fig:framework_agent}
  \vspace{-4mm}
\end{figure}

%% file: tables/main.tex
\begin{table*}[t]
\scriptsize
\centering
\newcolumntype{C}[1]{>{\centering\arraybackslash}p{#1}}
\newcolumntype{L}[1]{>{\arraybackslash}p{#1}}
\newcolumntype{R}{>{\raggedleft\arraybackslash}X}
\begin{tabularx}{\textwidth}{lC{6.4mm}C{6.4mm}C{6.4mm}C{6.4mm}C{6.4mm}C{6.4mm}C{6.4mm}C{6.4mm}C{6.4mm}C{6.4mm}C{6.4mm}C{6.4mm}C{6.4mm}C{6.4mm}}
\toprule
\multicolumn{1}{l|}{Method} & Depth & Spatial & Jigsaw & Vis corr. & Sem. Corr. & Art & Count & Fun. Corr. & Local. & Multi-view & Refl. & Fore. & IQ & Sim. \\ \midrule
\multicolumn{15}{c}{Open-source multimodal LLMs} \\ \midrule
\multicolumn{1}{l|}{OpenFlamingo-v2~\cite{awadalla2023openflamingo}} & \underline{54.0} & 43.4 & 47.3 & 25.6 & 30.2 & 52.1 & 21.7 & \textbf{36.2} & \underline{52.0} & 41.4 & 43.3 & 15.9 & 23.3 & \underline{55.2} \\
\multicolumn{1}{l|}{InstructBLIP-7B~\cite{instructblip}} & 51.6 & 56.6 & 52.7 & 30.8 & 30.9 & 47.9 & 29.2 & \underline{23.9} & 44.8 & \textbf{58.7} & 29.9 & \textbf{29.6} & 23.3 & 46.3 \\
\multicolumn{1}{l|}{InstructBLIP-13B~\cite{instructblip}} & 51.6 & 65.7 & 52.7 & 29.7 & \underline{32.4} & 50.4 & 30.8 &  22.3 & \underline{52.0} &  54.1 & \textbf{46.3} & 13.6 & 26.0 &  46.3 \\
\multicolumn{1}{l|}{CogVLM~\cite{wang2023cogvlm}} & 50.8 & 67.1 & 52.7 & 20.9 & 23.6 & 49.6 & 46.3 & \underline{23.9} & 43.2 & \underline{57.1} & 26.9 & 24.2 & 26.7 & 46.3 \\
\rowcolor{lightblue}
\multicolumn{1}{l|}{LLaVA-v1.5-7B~\cite{liu2023llava}} & 52.4 & 61.5 & 11.3 & 25.6 & 23.0 & 47.9 & 43.3 & 21.5 &  48.8 & 49.6 & 36.6 & \underline{28.0} & 24.0 & 46.3 \\
\rowcolor{lightgreen}
\multicolumn{1}{l|}{LLaVA-v1.5-13B~\cite{liu2023llava}} & 53.2 & 67.8 & \underline{58.0} & 29.1 & \underline{32.4} & 47.9 & \textbf{50.0} &  20.8 & \underline{47.2} & 41.4 & \underline{45.5} & 27.3 & \underline{28.0} & 46.3 \\ 
\midrule
\rowcolor{lightblue}
\multicolumn{1}{l|}{Ours (LLaVA-7B)} & 51.6 & \textbf{78.8} & 56.7 & \underline{33.1} & \underline{32.4} & \underline{54.7} & 41.2 & 21.5 & \textbf{56.6} & 55.6 & 37.0 & 26.5 & 23.3 & \textbf{58.5} \\
\rowcolor{lightgreen}
\multicolumn{1}{l|}{Ours (LLaVA-13B)} & \textbf{58.1} & \underline{69.9} & \textbf{64.0} & \textbf{34.3} & \textbf{34.5} & \textbf{58.1} & 47.2 & \underline{23.9} & 51.6 & 51.1 & 45.1 & 26.5 & \textbf{28.0} & 45.9 \\
\midrule
\multicolumn{15}{c}{API-based models} \\ 
\midrule
\multicolumn{1}{l|}{Qwen-VL-Max~\cite{bai2023qwen}} & 58.9 & 77.6 & 3.3 & 22.7 & 29.3 & 37.6 & 55.8 & 28.5 & 49.6 & 53.4 & \textbf{49.3} & 47.7 & 22.0 & 51.5 \\
\multicolumn{1}{l|}{Gemini Pro~\cite{gemini}} & 50.0 & 67.1 &  54.0 & 37.2 & 22.1 & 49.5 & \textbf{65.0} & 32.3 & 46.4 & 41.4 & \underline{46.3} & 45.5 & 27.3 & 55.9 \\
\multicolumn{1}{l|}{Claude 3 OPUS~\cite{claude3}} & 57.3 & 57.3 & 32.7 & 31.4 & 20.7 & 60.7 & 49.2 & 22.3 & 46.4 & 57.9 & 27.6 & 62.1 & 21.3 & 70.6 \\
\rowcolor{lightblue}
\multicolumn{1}{l|}{GPT-4o~\cite{gpt4}} & 74.2 & 69.2 & 55.3 & 75.0 & 54.0 & \underline{82.9} & 51.7 & 39.2 & 56.0 & \underline{60.2} & 38.8 & \textbf{85.6} & \textbf{30.0} & 65.4 \\ \rowcolor{lightorange}
\multicolumn{1}{l|}{GPT-4o (+ SoM + orig.)$^\dagger$} & 75.0 & 82.5 & - & - & - & - & - & - & - & - & - & - & - & - \\ \rowcolor{lightorange}
\multicolumn{1}{l|}{GPT-4o (+ Visprog)$^\dagger$} & 46.8 & 37.8 & - & - & - & - & - & - & - & - & - & - & - & - \\ 
\rowcolor{lightorange}
\multicolumn{1}{l|}{GPT-4o (+ Sketchpad)} & \textbf{83.9}$^\dagger$ & \underline{81.1}$^\dagger$ & \underline{70.7}$^\dagger$ & \underline{80.8}$^\dagger$ & \underline{58.3}$^\dagger$ & 77.19$^*$ & \underline{66.7}$^*$  & \underline{42.1}$^*$ & \textbf{65.4}$^*$ & 45.6$^*$ & 33.1$^*$ & 79.0$^*$ & 22.8$^*$ & \textbf{84.2}$^*$  \\ 
\midrule
\rowcolor{lightblue}
\multicolumn{1}{l|}{Ours (GPT-4o)} & \underline{80.3} & \textbf{81.8} & \textbf{75.3} & \textbf{85.5} & \textbf{58.3} & \textbf{83.0} & 61.7 & \textbf{55.4} & \underline{59.0} & \textbf{60.2} & 35.1 & \underline{84.8} & \underline{28.7} & \underline{75.3} \\ 
\bottomrule
\end{tabularx}
\vspace{-0.1in}
\caption{\textbf{Quantitative results. } Experimental results on the BLINK benchmark~\cite{fu2024blink}. $^\dagger$ denotes results from the previous work~\cite{hu2024visual}, and $^*$ represents results collected via official codebase. The best result is highlighted in \textbf{Bold} and the second underlined.}
\vspace{-0.17in}
\label{exp:blink}
\end{table*}

%% file: sec/4_exp.tex
\section{Experiments}
\label{sec:exp}

\input{tables/main_seed}

\paragraph{Datasets and Evaluation}

To verify the effectiveness of \ours, we conduct experiments on two benchmarks: BLINK~\cite{fu2024blink} and Seedbench~\cite{li2024seed}, and compare our model against previous works. These benchmarks contain various tasks covering visual perception and spatial understanding. \textbf{BLINK} includes 14 visual perception tasks with a total of 3,807 multiple-choice questions, while \textbf{SeedBench} covers 9 classical spatial understanding tasks with a total of 14k image-QA pairs, including scene understanding, instance interaction, and visual reasoning. 
There are some overlapping tasks between the two benchmarks; however, the main difference is that BLINK focuses on evaluating visual perception, where tasks are designed to be solvable by humans at a glance while hard to answer correctly for MLLMs. In contrast, SeedBench emphasizes models' visual spatial understanding, involving complex tasks with small objects or intricate descriptive prompts.
For evaluation, since the tasks in these datasets are single-choice questions, we follow their protocol by using GPT to map the open-form predictions from MLLMs to the fixed set of choices and perform string matching to report accuracy for each task.

\input{figures/demo}

\subsection{Quantitative Analysis}
In this subsection, we evaluate the effectiveness and performance of our \ours. Note that, to ensure a fair comparison with previous SoTA models, we fix the multimodal LLMs to the same ones used in the compared methods for quantitative evaluation. For vision models, we use exactly the same models as those employed in the prior work on Visual Sketchpad~\cite{hu2024visual}.


\vspace{0.1in}
\noindent
{\bf Can \ours~propose effective solutions?}
To verify this point, we conducted experiments on BLINK and SeedBench, reporting performance using three different multimodal LLMs ({\em i.e.}, LLaVA-7B, LLaVA-13B, and GPT-4o) as fixed MLLMs. The results are shown in Tables~\ref{exp:blink} and \ref{exp:seed}. Our method demonstrates its ability to propose useful solutions with either comparable or improved performance relative to its own base model. Notably, with the routing approach, very significant performance boosts are observed in certain tasks, such as function correspondence ($+15\%$ over GPT-4o) and jigsaw solving ($+20\%$ over GPT4o), spatial understanding ($+17\%$ over LLaVA-7B), and jigsaw again ($+6\%$ over LLaVA-13B). Consistent performance improvements are also seen on SeedBench, particularly for multi-instance understanding tasks like instance interaction and reasoning, with a $\approx 3\%$ increase, verifying the effectiveness of our proposed solution router.

\vspace{0.05in}
\noindent
{\bf Comparison with augmented frameworks for MLLMs}
We further compare our framework with other augmentation frameworks for MLLMs, such as SoM~\cite{yang2023set}, Visprog~\cite{gupta2023visual}, and Visual Sketchpad~\cite{hu2024visual}.
Visual Sketchpad~\cite{hu2024visual} allows LMs to adjust their solution based on intermediate visual results from other tools. To demonstrate that our solution proposer with multi-agent cooperation can produce better solution plans than Visual Sketchpad, we fixed the visual tools and the LM as used in their approach and reported the performance of our proposed solutions in Table~\ref{exp:blink}. Benefiting from multi-agent cooperation, our approach achieves comparable or better performance than the previous SoTA, highlighting the effectiveness of the solution proposer. Most importantly, our proposed solutions are general and not limited to specific samples within the task. As a result, we significantly reduce the API calling cost; see Figure~\ref{exp:time_cost} for more details. Last but not least, comparing with previous visual programming work of Visporg, we achieve $+\approx30\%$ over depth and spatial tasks, demonstrating our approach can propose a stronger pre-defined solution.


\subsection{Qualitative Analysis}
\label{sec:qualitative_analysis}
Fig.~\ref{fig:demo_large} shows qualitative examples of our proposed \ours. It samples a few examples from a given task, defined by the user’s constraints and task details (e.g., image and prompt), and passes them to ~\ours. The “solution proposer” then generates a pool of robust solutions for the task. Simultaneously, the “metric router” generates a performance curve showing the trade-off between time cost and accuracy based on selected metrics (\emph{e.g.} GPTScore). Unlike existing methods, our approach generates a solution pool from which users can choose the best option based on their constraints. Additionally, our framework provides solutions tailored to the entire task, rather than to individual samples. Additional examples are provided in supplement.

\input{tables/ablation_model}

\input{figures/exp_iteranalysis}
\vspace{-2mm}
\subsection{Model Analysis}
\vspace{-2mm}
\paragraph{Ablation studies of the multi-agent corporation.} 
In the solution router, we leverage multi-agent conversation to improve the quality and robustness of the generated solutions. We conduct ablation studies on the multi-agent component of the proposer to verify this, with the results shown in~\cref{exp:ablation}. Specifically, we run the solution router on the first five tasks (listed in~\cref{exp:blink}) in the BLINK dataset, with three runs per task, each allowing a max of six conversation iterations. Without the code debugger, the code checker cannot access the intermediate results of the solution, resulting in a significantly performance accuracy drop of $10\%$. Without the code checker, there is no feedback on execution results, which not only reduces the performance but also substantially increases the error rate during solution execution. Furthermore, after ablating the requirement checker, we observe both performance and solution correctness degrade compared to the full model. Lastly, without the repetition checker, the average number of proposed solutions decreases significantly by $33\%$, verifying the effectiveness of the repetition checker in enhancing solution diversity.

\vspace{0.05in}
\noindent
{\bf Routing time and API calling cost.} 
In our solution router, agents iteratively converse to generate the final solutions. As the number of existing solutions in the pool grows, the router may take more time to propose a novel solution. Therefore, we further investigate the routing time cost with varying numbers of solutions in the pool. The average time cost per solution and per iteration is reported in~\cref{exp:time_cost} (top). We observe that the time cost per solution increases as the number of existing solutions grows. We assume this is due to the increasing complexity of the task, requiring the router to utilize the maximum number of iterations to derive the final solution. On average, it takes approximately 8 minutes to generate a solution. Notably, since the generated solutions are applicable to all samples within a task, we only need to perform solution routing once per task, rather than for each sample. We compare execution and routing costs with Visual Sketchpad in~\cref{exp:time_cost} (bottom). Execution cost refers to the time from input prompt to final answer, while routing cost is the time spent coordinating tools (execution time minus tool-calling time). One can find that with the pre-planned solutions, our execution cost is lower. Additionally, as the proposed solutions are reusable across all task instances, routing cost per sample is nearly zero, significantly less than the on-line routing in previous work.

\input{tables/ablation_cost}
\input{tables/ablation_apicost}
Furthermore, as we use a GPT model for the solution router, we report the average API cost and compare it with previous work, Visual Sketchpad~\cite{hu2024visual} (see~\cref{exp:api_cost}). A key benefit of our approach is that we perform routing only for a few runs, with the produced solution applicable to all samples, significantly reducing the cost. In contrast, Sketchpad requires an API call for every sample, resulting in almost five times the cost of our approach on the BLINK dataset.

\vspace{0.1in}
\noindent
{\bf Best answer happen in which run progressive performance analysis.} 
In the solution router, we set a maximum number of conversation iterations for the multi-agent cooperation. As mentioned in previous studies~\cite{du2023improving}, multi-agent conversation or debate can lead to error propagation, reducing performance after multiple iterations. To investigate this, we conducted experiments to analyze performance as the number of iterations increased, with results shown in~\cref{exp:iter_analysis}. Specifically, we randomly selected 10 tasks from the BLINK dataset and ran our solution router to generate solutions, setting the maximum number of iterations to six. As shown in the figure, we observe that solutions with the best performance occur around $2–4$ iterations. 

%% file: tables/main_seed.tex
\begin{table}[t]
\scriptsize
\centering
\newcolumntype{C}[1]{>{\centering\arraybackslash}p{#1}}
\newcolumntype{L}[1]{>{\arraybackslash}p{#1}}
\newcolumntype{R}{>{\raggedleft\arraybackslash}X}
\begin{tabularx}{0.97\columnwidth}{lC{7.7mm}C{7.7mm}C{7.7mm}C{7.7mm}C{7.7mm}}
\toprule
\textbf{Model} & \textbf{Avg.} & \textbf{Scene} & \textbf{Id} & \textbf{Attri.} & \textbf{Locat.} \\
\midrule
InstructBLIP~\cite{instructblip} & 51.5 & 58.9 & 49.7 & 61.7 & 35.1 \\
LLaVA-v1.5-7B~\cite{liu2023llava} & 57.7 & 63.7 & 62.4 & 66.7 & 51.3 \\
MiniGPT-4~\cite{zhu2023minigpt} & 45.9 & 56.3 & 49.2 & 45.8 & 37.9 \\
OpenFlamingo~\cite{awadalla2023openflamingo} & 36.1 & 46.7 & 42.3 & 31.7 & 33.4 \\
Qwen-VL-Chat~\cite{bai2023qwen} & 50.9 & 56.5 & 47.6 & 54.8 & 46.9 \\
CogVLM~\cite{wang2023cogvlm} & 42.4 & 51.7 & 43.5 & 38.9 & 33.8 \\
InternLM~\cite{zhang2023internlm} & 69.2 & 77.5 & 73.5 & 74.8 & 65.4 \\
\rowcolor{lightblue}
GPT-4o~\cite{gpt4} & 75.6 & 77.3 & \textbf{79.7} & 79.2 & \textbf{71.0} \\
\midrule
\rowcolor{lightblue}
Ours (GPT-4o) & \textbf{75.8} & \textbf{78.3} & 78.3 & \textbf{79.7} & 70.1 \\ 
\midrule
\textbf{Model} & \textbf{Count.} & \textbf{Spatial} & \textbf{Inter.} & \textbf{Reason.} & \textbf{Text}  \\
\midrule
InstructBLIP~\cite{instructblip} & 58.1 & 34.9 & 47.4 & 55.9 & 61.4 \\
LLaVA-v1.5-7B~\cite{liu2023llava} & 60.2 & 38.5 & 47.4 & 59.8 & 69.0 \\
MiniGPT-4~\cite{zhu2023minigpt} & 45.3 & 32.6 & 47.4 & 57.1 & 41.8 \\
OpenFlamingo~\cite{awadalla2023openflamingo} & 27.4 & 29.8 & 29.9 & 47.7 & 35.6 \\
Qwen-VL-Chat~\cite{bai2023qwen} & 54.2 & 40.3 & 55.7 & 55.0 & 47.4 \\
CogVLM~\cite{wang2023cogvlm} & 29.4 & 33.6 & 45.4 & 53.5 & 51.5 \\
InternLM~\cite{zhang2023internlm} & 65.8 & 57.5 & 71.1 & 75.8 & 61.2 \\
\rowcolor{lightblue}
GPT-4o~\cite{gpt4} & \textbf{68.1} & \textbf{63.8} & 78.6 & 81.2 & 69.8 \\
\midrule
\rowcolor{lightblue}
Ours (GPT-4o) & 67.7 & 62.8 & \textbf{80.6} & \textbf{84.5} & \textbf{69.9} \\
\bottomrule
\end{tabularx}
\vspace{-0.1in}
\caption{\textbf{Quantitative results on Seedbench}~\cite{li2024seed}.}
\vspace{-0.25in}
\label{exp:seed}
\end{table}


%% file: figures/demo.tex
\begin{figure*}[t]
  \centering
  \includegraphics[page=1, trim={45 90 20 65}, clip, width=0.75\textwidth]{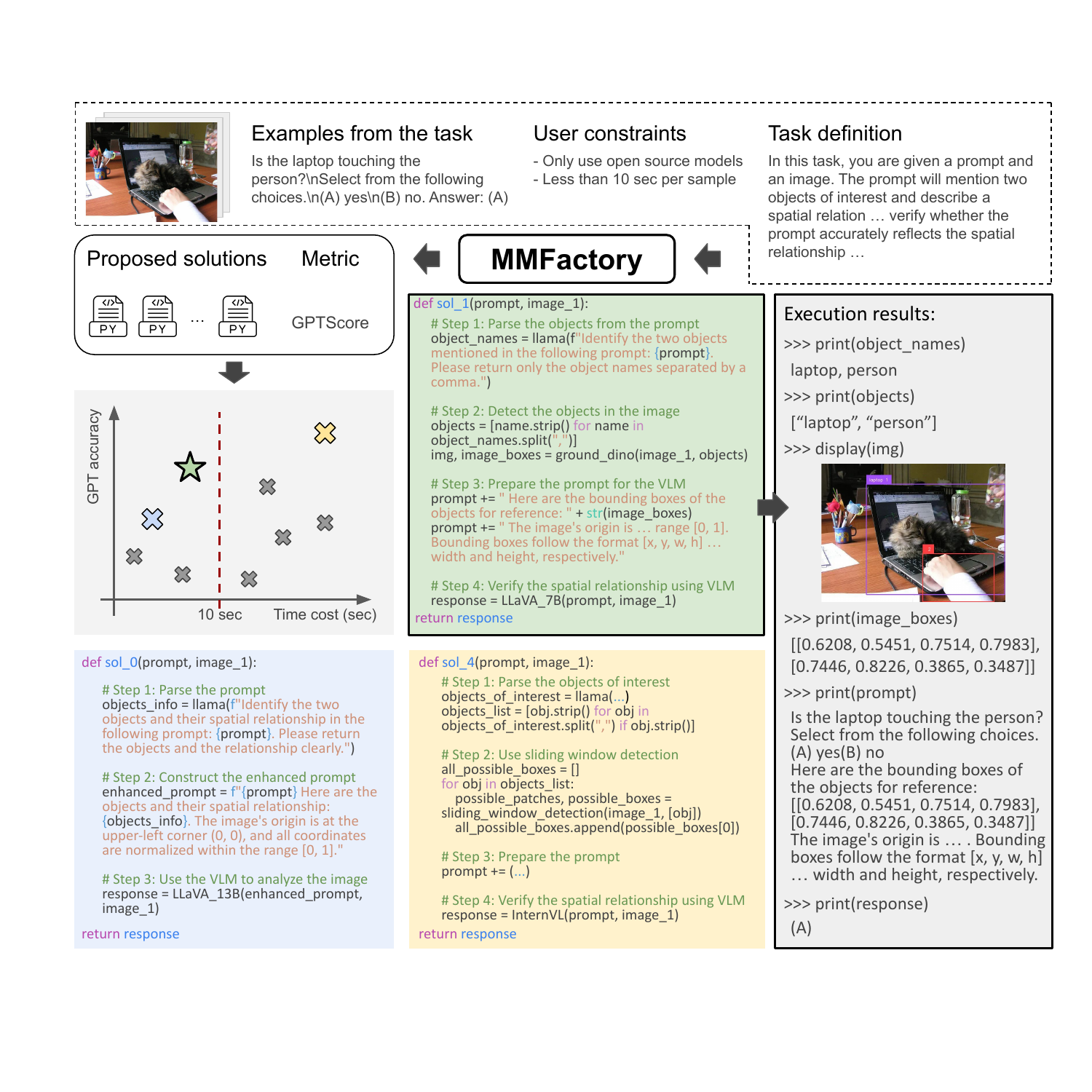}
  \vspace{-3mm}
  \caption{\textbf{Qualitative examples of \ours}. \ours~showcases its abilities to use and combine models by automatically constructing better prompts for MLLMs (in \textcolor{blue}{Sol 0}) and developing solutions with similar logic but utilizing stronger models (in \textcolor{yellow}{Sol 4}).}
  \label{fig:demo_large}
  \vspace{-5mm}
\end{figure*}

%% file: tables/ablation_model.tex
\begin{table}[]
\small
\centering
\resizebox{0.9\columnwidth}{!}{
\begin{tabular}{cccc}
\toprule
\multicolumn{1}{c|}{Model} & Acc & Error rate & Avg. \# sols \\ \hline
\multicolumn{1}{l|}{Full model} & \textbf{50.5} & \textbf{0.0} & \textbf{3.0} \\ 
\multicolumn{1}{l|}{\hspace{3mm} (-) \textit{code debugger}} & 40.0 & 1.7 & 2.8 \\
\multicolumn{1}{l|}{\hspace{3mm} (-) \textit{code checker}} & 33.3 & 20.8 & \textbf{3.0} \\ 
\multicolumn{1}{l|}{\hspace{3mm} (-) \textit{requirement checker}} & 48.1 & 0.5 & 2.4 \\ 
\multicolumn{1}{l|}{\hspace{3mm} (-) \textit{repetition checker}} &  40.5 & 17.8 & 2.0 \\ \bottomrule
\end{tabular}}
\vspace{-0.1in}
\caption{\textbf{Ablation.} of significance of multi-agent conversation.}
\vspace{-0.26in}
\label{exp:ablation}
\end{table}

%% file: figures/exp_iteranalysis.tex

\begin{figure}[t]
  \centering
  \includegraphics[page=20, trim={200 85 190 90}, clip, width=0.8\columnwidth]{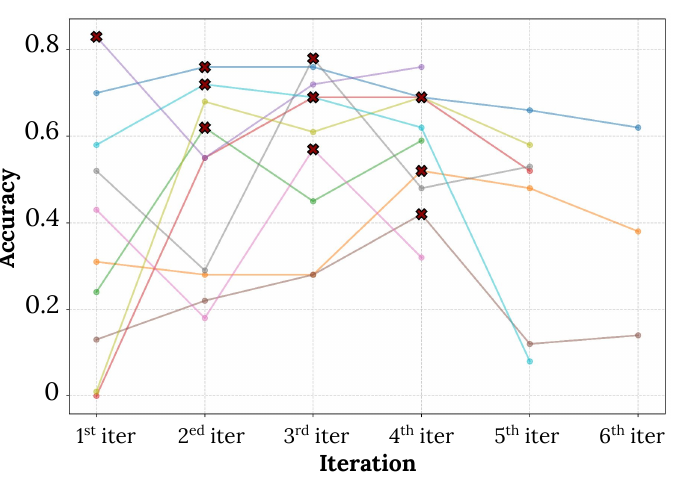}
  \vspace{-3mm}
  \caption{\textbf{Ablation.} Performance analysis with iteration. Lines in different colors represent different runs. Red cross denotes the highest performance in the run.}
  \vspace{-0.1in}
  \label{exp:iter_analysis}
  \vspace{-4mm}
\end{figure}

%% file: tables/ablation_cost.tex
\begin{figure}[t]
  \centering
  \begin{subfigure}{\columnwidth}
    \centering
    \includegraphics[page=18, trim={180 135 130 58}, clip, width=0.80\columnwidth]{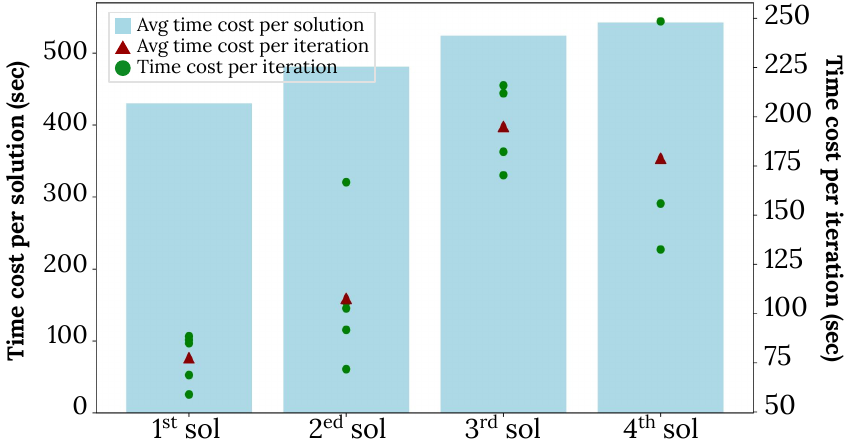}
    \label{exp:time_cost_a}
    \vspace{1mm}
  \end{subfigure}

  \begin{subfigure}{\columnwidth}
    \centering
    \small
    \begin{tabular}{l|cc|cc}
      \toprule
      \multicolumn{1}{c|}{\multirow{2}{*}{Model}} & \multicolumn{2}{c|}{Execution cost (sec)} & \multicolumn{2}{c}{Routing cost (sec)} \\ \cline{2-5} 
      \multicolumn{1}{c|}{} & Mean & Variance & Mean & Variance \\ \hline
      Sketchpad \cite{hu2024visual} & 19.96 & 43.86 & 18.20 & 30.90 \\ \hline
      Ours & 9.74 & 29.43 & $\approx$ 0.00 & $\approx$ 0.00 \\
      \bottomrule
    \end{tabular}
    \label{exp:time_cost_b}
  \end{subfigure}
  \vspace{-3mm}
  \caption{{\bf Computational time.} Solution generation cost plot (top). Average execution and routing cost \textbf{per sample} (bottom).}
  \label{exp:time_cost}
  \vspace{-4mm}
\end{figure}

%% file: tables/ablation_apicost.tex
\begin{table}[]
\small
\resizebox{\columnwidth}{!}{
\begin{tabular}{cccccc}
\toprule
\multicolumn{1}{c|}{Model} & Depth & Spatial & Jigsaw & \begin{tabular}[c]{@{}c@{}}Vis.\\ Corr.\end{tabular} & \begin{tabular}[c]{@{}c@{}}Sem.\\ Corr.\end{tabular} \\ \hline
\multicolumn{1}{c|}{Sketchpad \cite{hu2024visual}} & 0.211 & 0.232 & 0.224 & 0.281 & 0.230 \\
\multicolumn{1}{l|}{Ours} & 0.064 & 0.045 & 0.041 & 0.034 & 0.058 \\ 
\bottomrule
\end{tabular}}
\vspace{-0.1in}
\caption{\textbf{API calling cost analysis} per 10 samples (in USD).}
\vspace{-0.25in}
\label{exp:api_cost}
\end{table}

%% file: sec/5_conculsion.tex
\vspace{-3mm}
\section{Conclusion}
\vspace{-1mm}
\label{sec:conclusion}
Selecting the right multimodal LLM for a task can be difficult, especially without domain-specific knowledge or clear user requirements. In this paper, we present a framework to help users select the most suitable solution from a solution pool for a given tasks based on their specific constraints. Our approach uses a multi-agent debate mechanism to generate robust and well-reasoned solution. Unlike sample-specific solutions, our framework provides guidance that applies broadly across all examples for a given task. Through extensive experiments, we demonstrate that our method outperforms current state-of-the-art approaches.

\section*{Acknowledgements}
{\small This work was funded, in part, by the Vector Institute for AI, Canada CIFAR AI Chairs, NSERC Canada Research Chair (CRC), and NSERC Discovery and Discovery Accelerator Supplement Grants. Resources used in preparing this research were provided, in part, by the Province of Ontario, the Government of Canada through CIFAR, the Digital Research Alliance of Canada\footnote{\url{alliance.can.ca}}, companies\footnote{\url{https://vectorinstitute.ai/\#partners}} sponsoring the Vector Institute, and Advanced Research Computing at the University of British Columbia. Additional hardware support was provided by John R. Evans Leaders Fund CFI grant and Compute Canada under the Resource Allocation Competition award.}